\documentclass[11pt,a4paper]{article}
\usepackage[hyperref]{acl2018}
\usepackage{array}
\usepackage{times}
\usepackage{latexsym}
\usepackage{amsmath}
\usepackage{bbm}
\usepackage{booktabs}
\usepackage{arydshln}
\usepackage{url}
\usepackage{natbib}
\usepackage{graphicx}
\usepackage{subcaption}
\usepackage{tikz}
\usetikzlibrary{shapes,arrows,shadows}
\pgfdeclarelayer{background}
\pgfdeclarelayer{foreground}
\pgfsetlayers{background,main,foreground}
\usepackage[linesnumbered,ruled]{algorithm2e}
\newcommand{\tabitem}{~~\llap{\textbullet}~~}

\DeclareMathOperator*{\argmax}{argmax}

\aclfinalcopy 

\title{Distilling Knowledge for Search-based Structured Prediction}

\author{Yijia Liu, Wanxiang Che\thanks{* Email corresponding.}, Huaipeng Zhao, Bing Qin, Ting Liu \\
	Research Center for Social Computing and Information Retrieval \\
	Harbin Institute of Technology, China \\
	{\tt \{yjliu,car,hpzhao,qinb,tliu\}@ir.hit.edu.cn}	
}
\date{}

\begin{document}
\maketitle
\begin{abstract}
	Many natural language processing tasks can be modeled into 
	structured prediction and solved as a  search problem.
	In this paper, we distill an ensemble of multiple
	models trained with different initialization 
	into a single model. 
	In addition to learning to
	match the ensemble's probability output on the reference states,
	we also
	use the ensemble to explore the search space and learn from
	the encountered states in the exploration. 
	Experimental results on
	two typical search-based structured prediction tasks --
	transition-based dependency parsing and neural machine translation
    show that
	distillation can effectively improve the single model's performance
	and 
	the final model achieves improvements of 1.32 in LAS and 2.65 in BLEU score
	on these two tasks respectively over
	strong baselines and it outperforms the greedy structured prediction models
	in previous literatures.
\end{abstract}

\section{Introduction}

Search-based structured prediction models the generation of
natural language structure (part-of-speech tags, syntax tree, translations, semantic graphs, etc.) 
as a search problem \cite{collins-roark:2004:ACL,
	liang-EtAl:2006:COLACL,
	zhang-clark:2008:EMNLP,
	huang-fayong-guo:2012:NAACL-HLT,
	NIPS2014_5346,
	goodman-vlachos-naradowsky:2016:P16-1}.
It has drawn a lot of research attention in recent years
thanks to its competitive performance on both accuracy and running time.
A stochastic policy that controls the whole search process is
usually learned by imitating a \textit{reference policy}.
The imitation is usually addressed as training a classifier to predict
the reference policy's search action on the encountered states when
performing the reference policy.
Such imitation process can sometimes be problematic. 
One problem is the ambiguities of the reference policy, in which
multiple actions lead to the optimal structure but usually, only one
is chosen as training instance \cite{goldberg-nivre:2012:PAPERS}.
Another problem is the discrepancy between training and testing, in which
during the test phase, the learned policy enters non-optimal
states whose search action is never learned \cite{pmlr-v9-ross10a,pmlr-v15-ross11a}.
All these problems harm the generalization ability of search-based
structured prediction and lead to poor performance.

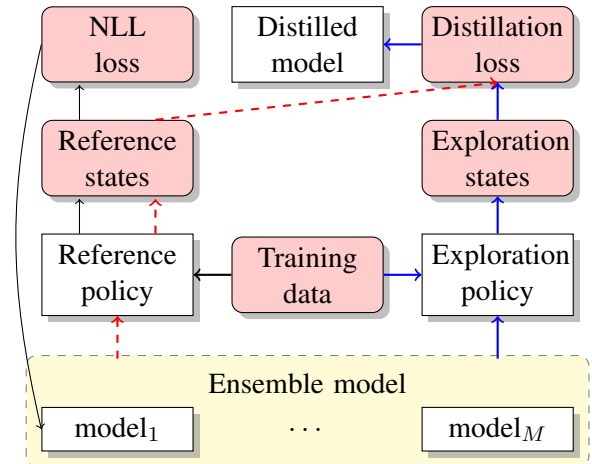
\begin{figure}[t]
\tikzstyle{ann} = [above, text width=5em, text centered]
\tikzstyle{data} = [draw, minimum width=3em, fill=red!20, minimum height=1.5em, text centered, text width=4.5em, rounded corners, drop shadow]
\tikzstyle{model}=[draw, fill=white!20, minimum width=4.5em, minimum height=1.5em, text width=4.5em, text centered, drop shadow]
\centering
\begin{tikzpicture}
\node (data) [data]  {Training data};
\path (data.west)+(-1.5,0) node (refp)[model] {Reference policy};
\path (data.east)+(1.5,0) node (expp)[model] {Exploration policy};
\path (refp.north)+(0,1) node (refs)[data] {Reference states};
\path (refs.north)+(0,1) node (reft)[data] {NLL\\loss};
\path (expp.north)+(0,1) node (exps)[data] {Exploration states};
\path (exps.north)+(0,1) node (expt)[data] {Distillation loss};
\path (expt.west)+(-1.5,0) node (distill)[model] {Distilled model};
\path (refp.south)+(0,-1.5) node (m1)[model] {model$_1$};
\path (data.south)+(0,-1.2) node (ens)[above, text width=10em, text centered] {Ensemble model}; 
\path (data.south)+(0,-1.7) node (dots)[ann] {$\dots$};
\path (expp.south)+(0,-1.5) node (mm)[model] {model$_M$};

\path (refp.south)+(0,-0.7) node (ens14) {};
\path (expp.south)+(0,-0.7) node (ens34) {};
\path (refs.south)+(-0.5,0.1) node (refssouth14) {};
\path (refs.south)+(+0.5,0.1) node (refssouth34) {};
\path (reft.south)+(-0.5,0.1) node (reftsouth14) {};
\path (reft.south)+(+0.5,0.1) node (reftsouth34) {};

\draw[->, thick] (data.west) to (refp.east);
\draw[->] (refp.north)+(-0.5,0) to (refssouth14);
\draw[->] (refs.north)+(-0.5,0) to (reftsouth14);
\draw[->] (reft.west) to [bend right=13] (m1.west);

\draw[->, color=red, dashed, thick] (refp.north)+(+0.5,0) to (refssouth34);
\draw[->, color=red, dashed, thick] (refs.north)+(+0.5,0) to (expt.south);
\draw[->, color=red, dashed, thick] (ens14) to (refp.south);

\draw [->, color=blue, thick] (data.east) to (expp.west);
\draw [->, color=blue, thick] (expp.north) to (exps.south);
\draw [->, color=blue, thick] (exps.north) to (expt.south);
\draw [->, color=blue, thick] (expt.west) to (distill.east);
\draw [->, color=blue, thick] (ens34) to (expp.south);

\begin{pgfonlayer}{background}
\path (m1.west |- m1.south)+(-0.2,-0.2) node (a) {};
\path (mm.east |- mm.south)+(+0.2,-0.2) node (b) {};
\path (ens.north -| mm.east)+(+0.2,+0.1) node (c) {};
\path (m1.west |- ens.north)+(-0.2,+0.1) node (d) {};
\path[fill=yellow!20, rounded corners, draw=black!50, dashed] (a) rectangle (c);
\end{pgfonlayer}
\end{tikzpicture}
	\caption{Workflow of our knowledge distillation for search-based
		structured prediction. The yellow bracket represents the ensemble
		of multiple models trained with different initialization. 
		The dashed red line shows our {\it distillation from reference} (\S\ref{sec:distill_ref}).
		The solid blue line shows our {\it distillation from exploration} (\S\ref{sec:distill_explore}).}\label{fig:workflow}
\end{figure}

\begin{table*}[t]
	\centering
	\small
	\begin{tabular}{rp{0.43\textwidth}p{0.43\textwidth}}
		\hline
		& Dependency parsing  & Neural machine translation\\[0.1em]
		\hline
		$s_t$ & $(\sigma, \beta, A)$, where $\sigma$ is a stack, $\beta$ is a buffer, and $A$ is the partially generated tree & $(\text{\$}, y_1, y_2, ..., y_t)$, where \$ is the start symbol. \\[0.1em]
		$\mathcal{A}$ & \{{\sc Shift}, {\sc Left}, {\sc Right}\} & pick one word $w$ from the target side vocabulary $\mathcal{W}$. \\[0.1em]
		$\mathcal{S}_0$ & \{$([\ ], [1, .., n], \emptyset)$\} & \{$(\text{\$)} $\}\\[0.1em]
		$\mathcal{S}_T$ & $\{([\text{ROOT}], [\ ], A)\}$ & $\{(\text{\$}, y_1, y_2, ..., y_m)\}$ \\[0.1em]
		$\mathcal{T}(s, a)$ & \tabitem {\sc Shift}: $(\sigma, j | \beta) \to (\sigma | j, \beta)$ & $(\text{\$}, y_1, y_2, ..., y_t) \to (\text{\$}, y_1, y_2, ..., y_t, y_{t+1}=w)$ \\[0.1em]
		& \tabitem {\sc Left}: $(\sigma |  i\ j, \beta) \to (\sigma | j, \beta)\quad A \gets A \cup \{i \leftarrow j\}$ & \\[0.1em]
		& \tabitem {\sc Right}: $(\sigma |  i\ j, \beta) \to (\sigma | i, \beta)\quad A \gets A \cup \{i \rightarrow j\}$ & \\
		\hline
	\end{tabular}
	\caption{The search-based structured prediction view of
		transition-based dependency parsing \citep{nivre2008algorithms} 
		and neural machine translation \citep{NIPS2014_5346}.}\label{tbl:search-nlp}
\end{table*}

Previous works tackle these problems from two directions.
To overcome the ambiguities in data, techniques like {\it ensemble} are often adopted \cite{Dietterich2000}.
To mitigate the discrepancy, exploration is encouraged during the training process \cite{
	pmlr-v9-ross10a,
	pmlr-v15-ross11a,
	goldberg-nivre:2012:PAPERS,
	NIPS2015_5956,
	goodman-vlachos-naradowsky:2016:P16-1}.
In this paper, we propose to consider these two problems in an integrated
{\it knowledge distillation} manner \cite{DBLP:journals/corr/HintonVD15}.
We distill a single model from the ensemble of several baselines trained with different initialization 
by matching the ensemble's output distribution on the reference states.
We also let the ensemble randomly explore the search space and learn the single
model to mimic ensemble's distribution on the encountered exploration states.
Combing the distillation from reference and exploration further improves
our single model's performance.
The workflow of our method is shown in Figure \ref{fig:workflow}.

We conduct experiments on two typical search-based structured prediction tasks: 
transition-based dependency parsing
and neural machine translation. The results of both these two experiments 
show the effectiveness
of our knowledge distillation method by outperforming strong baselines.
In the parsing experiments, an improvement of 1.32 in LAS is achieved and
in the machine translation experiments, such improvement is 2.65 in BLEU.
Our model also outperforms the greedy models in previous works.

Major contributions of this paper include:
\begin{itemize}
	\item We study the knowledge distillation in search-based structured prediction
	and propose to distill the knowledge of an ensemble into a single model by
	learning to match its distribution on both the reference states (\S\ref{sec:distill_ref}) and exploration
	states encountered when using the ensemble to explore the search space (\S\ref{sec:distill_explore}).
	A further combination of these two methods is also proposed to improve the
	performance (\S\ref{sec:distill_both}).
	
	\item We conduct experiments on two search-based structured prediction problems:
	transition-based dependency parsing and neural machine translation.
	In both these two problems, the distilled model significantly improves over strong
	baselines and outperforms other greedy structured prediction (\S\ref{sec:exp-res}).
	Comprehensive analysis empirically shows the feasibility of our distillation method (\S\ref{sec:analysis}).
\end{itemize}

\section{Background}

\subsection{Search-based Structured Prediction}\label{sec:sbsp}

Structured prediction maps an input $\mathbf{x}=(x_1, x_2, ..., x_n)$ to its structural output
$\mathbf{y}=(y_1, y_2, ..., y_m)$, where each component of $\mathbf{y}$ has some
internal dependencies. Search-based structured prediction \citep{collins-roark:2004:ACL,
	daume05search,
	Daume:2009:SSP:1541660.1541689,
	pmlr-v9-ross10a,
	pmlr-v15-ross11a,
	DBLP:journals/jair/DoppaFT14,
	TACL431,
	CKADL15}
models the generation of the structure as a search problem and it can be formalized
as a tuple $(\mathcal{S}, \mathcal{A}, \mathcal{T}(s, a), \mathcal{S}_0, \mathcal{S}_T)$, in which
$\mathcal{S}$ is a set of states,
$\mathcal{A}$ is a set of actions,
$\mathcal{T}$ is a function that maps $\mathcal{S}\times\mathcal{A} \to \mathcal{S}$,
$\mathcal{S}_0$ is a set of initial states, and
$\mathcal{S}_T$ is a set of terminal states.
Starting from an initial state $s_0\in \mathcal{S}_0$,
the structured prediction model repeatably chooses an action
$a_t \in \mathcal{A}$ by following a {\it policy} $\pi(s)$ and applies $a_t$ to $s_t$
and enter a new state $s_{t+1}$ as $s_{t+1} \gets \mathcal{T}(s_t, a_t)$,
until a final state $s_T \in \mathcal{S}_T$ is achieved.
Several natural
language structured prediction problems can be modeled under the search-based framework
including dependency parsing \citep{nivre2008algorithms} 
and neural machine translation \citep{liang-EtAl:2006:COLACL,NIPS2014_5346}.
Table \ref{tbl:search-nlp} shows the search-based structured prediction view of these two problems.

In the data-driven settings, $\pi(s)$ controls the whole search process and is usually
parameterized by a classifier $p(a \mid s)$ which outputs the probability of choosing
an action $a$ on the given state $s$. The commonly adopted greedy
policy can be formalized as choosing the most probable action with
$\pi(s) = \argmax_a p(a \mid s)$ at test stage. To learn an optimal
classifier, search-based structured prediction requires constructing a reference
policy $\pi_\mathcal{R}(s, \mathbf{y})$, which takes an input state $s$,
gold structure $\mathbf{y}$ and outputs its reference action $a$, and 
training $p(a\mid s)$ to imitate the reference policy.
Algorithm \ref{algo:generic} shows the common practices in training $p(a \mid s)$,
which involves: first, 
using $\pi_\mathcal{R}(s,  \mathbf{y})$ to generate a sequence of reference states and 
actions on the training data 
(line \ref{algo:generic:gen_start} to line \ref{algo:generic:gen_end} in Algorithm \ref{algo:generic});
second, using the states and actions on the reference sequences as examples
to train $p(a \mid s)$ with negative log-likelihood (NLL) loss
(line \ref{algo:generic:optim} in Algorithm \ref{algo:generic}),
\[
\mathcal{L}_{NLL} =  \sum_{s \in D} \sum_{a} -\mathbbm{1}\{a=\pi_\mathcal{R}\} \cdot \log p(a \mid s)
\]
where $D$ is a set of training data.

\begin{algorithm}[t]
	\KwIn{training data: $\{\mathbf{x}^{(n)}, \mathbf{y}^{(n)}\}_{n=1}^N$;
		the reference policy: $\pi_\mathcal{R}(s, \mathbf{y})$.}
	\KwOut{classifier $p(a|s)$.}
	$D \gets \emptyset$\; \label{algo:generic:gen_start}
	\For{$n \gets 1 ... N$}{
		$t \gets 0$\;
		$s_t \gets s_0(\mathbf{x}^{(n)})$\;
		\While{$s_t \notin \mathcal{S}_T$}{
			$a_t \gets \pi_\mathcal{R}(s_t, \mathbf{y}^{(n)})$\;
			$D \gets D \cup \{s_t\}$\;
			$s_{t+1}\gets \mathcal{T}(s_t, a_t)$\;
			$t \gets t + 1$\;
		}
	} \label{algo:generic:gen_end}
	optimize \(\mathcal{L}_{NLL}\)\;
	\label{algo:generic:optim}
	\caption{Generic learning algorithm for search-based structured prediction.
	}\label{algo:generic}
\end{algorithm}

The reference policy is sometimes sub-optimal and ambiguous which
means on one state, there
can be more than one action that leads to the optimal prediction.
In transition-based dependency parsing, \citet{goldberg-nivre:2012:PAPERS}
showed that
one dependency tree can be reached by several search sequences
using \citet{nivre2008algorithms}'s {\it arc-standard} algorithm.
In machine translation, the ambiguity problem also exists because one
source language sentence usually has multiple semantically correct translations
but only one reference translation is presented.
Similar problems have also been observed in semantic parsing
\cite{goodman-vlachos-naradowsky:2016:P16-1}.
According to \citet{Frnay2014ClassificationIT}, the widely used
NLL loss is vulnerable to ambiguous data which make it worse
for search-based structured prediction.

Besides the ambiguity problem, training and testing discrepancy is another
problem that lags the search-based structured prediction performance.
Since the
training process
imitates the reference policy, all the states in the training data are optimal
which means they are guaranteed to reach the optimal structure.
But during the test phase,
the model can predict non-optimal  states whose search action is never
learned. The greedy search which is prone to error propagation also
worsens this problem.

\subsection{Knowledge Distillation}
A cumbersome model, which could be
an ensemble of several models or a single model with larger number of
parameters, usually provides better generalization ability.
{\it Knowledge distillation} \citep{Bucilua:2006:MC:1150402.1150464,NIPS2014_5484,DBLP:journals/corr/HintonVD15}
is a class of methods for transferring the generalization ability of 
the  cumbersome {\it teacher model}  into a small  {\it student model}.
Instead of optimizing NLL loss,
knowledge distillation uses the distribution $q(y \mid x)$ outputted by the teacher model
as ``soft target'' and optimizes the knowledge distillation loss,
\[
\mathcal{L}_{KD} =  \sum_{x \in D} \sum_{y} -q(y
	\mid x) \cdot \log p(y \mid x).
\]
In search-based structured prediction scenario,
$x$ corresponds to the state $s$ and $y$ corresponds to the action $a$.
Through optimizing the distillation loss,
knowledge of the teacher model is learned by
the student model $p(y \mid x)$. When correct label is presented, NLL loss can be
combined with the distillation loss via simple interpolation as
\begin{align}\label{eq:distill}
\mathcal{L} = \alpha \mathcal{L}_{KD} + (1 - \alpha) \mathcal{L}_{NLL}
\end{align}


\section{Knowledge Distillation for Search-based Structured Prediction}

\subsection{Ensemble}
As \citet{DBLP:journals/corr/HintonVD15} pointed out, although the real objective
of a machine learning algorithm is to generalize well to new data, models are
usually trained to optimize the performance on training data, which bias the
model to the training data. 
In search-based structured prediction, such biases can result from
either the ambiguities in the training
data or the discrepancy between training and testing.
It would be more problematic to train $p(a \mid s)$
using the loss which is in-robust to ambiguities and
only considering the optimal states.

The effect of ensemble on ambiguous data has been studied in \citet{Dietterich2000}.
They empirically showed that ensemble can overcome the ambiguities in the training data.
\citet{daume05search} also use weighted ensemble of parameters from different
iterations as their final structure prediction model. 
In this paper, we consider to use ensemble technique to improve the
generalization ability of our search-based structured prediction model
following these works.
In practice, we train $M$ search-based 
structured prediction models with different initialized weights 
and ensemble them by the average
of their output distribution as $q(a \mid s) = \frac{1}{M} \sum_m q_m(a\mid s)$.
In Section \ref{sec:ens-on-states}, we empirically show that
the ensemble has the ability to choose a good search action
in the optimal-yet-ambiguous states and the non-optimal states.

\subsection{Distillation from Reference}\label{sec:distill_ref}

\begin{algorithm}[t]
	\KwIn{training data: $\{\mathbf{x}^{(n)}, \mathbf{y}^{(n)}\}_{n=1}^N$;
		the reference policy: $\pi_\mathcal{R}(s, \mathbf{y})$;
		the exploration policy: $\pi_\mathcal{E}(s)$ 
		which samples an action from the annealed ensemble $q(a\mid s)^{\frac{1}{T}}$}
	\KwOut{classifier $p(a\mid s)$.}
		$D \gets \emptyset$\; 
	\For{$n \gets 1 ... N$}{
		$t \gets 0$\;
		$s_t \gets s_0(\mathbf{x}^{(n)})$\;
		\While{$s_t \notin \mathcal{S}_T$}{
			\eIf{distilling from reference}{
				$a_t \gets \pi_\mathcal{R}(s_t, \mathbf{y}^{(n)})$\;
			}{
				$a_t \gets \pi_\mathcal{E}(s_t)$\;
			}
			$D \gets D \cup \{s_t\}$\;
			$s_{t+1}\gets \mathcal{T}(s_t, a_t)$\;
			$t \gets t + 1$\;
		}
	}
	\eIf{distilling from reference}{
	  	optimize $\alpha \mathcal{L}_{KD} + (1 - \alpha) \mathcal{L}_{NLL}$\;
	}{
		optimize $\mathcal{L}_{KD}$\;
	}
	\caption{Knowledge distillation for search-based structured prediction.}\label{algo:distill_ref}
\end{algorithm}

As we can see in Section \ref{sec:vani-exp}, ensemble indeed improves the
performance of baseline models.
However, real world deployment is usually constrained by computation
and memory resources. 
Ensemble requires running the structured prediction models for multiple times,
and that makes it less applicable in real-world problem.
To take the advantage of the
ensemble model while avoid running the models multiple times,
we use the knowledge distillation technique to distill
a single model from the ensemble.
We started from changing the NLL learning objective
in Algorithm \ref{algo:generic} 
into the distillation loss (Equation \ref{eq:distill})
as shown in Algorithm \ref{algo:distill_ref}.
Since such method learns the model on the states
produced by the reference policy, we name it as {\it distillation from reference}.
Blocks connected by in dashed red lines in Figure \ref{fig:workflow} show
the workflow of our {\it distillation from reference}.


\subsection{Distillation from Exploration}\label{sec:distill_explore}

In the scenario of search-based structured prediction, transferring the
teacher model's generalization ability into a student model 
not only includes
matching the teacher model's soft targets on the reference search sequence,
but also imitating the search decisions made by the teacher model.
One way to accomplish the imitation can be sampling search sequence
from the ensemble and learn from the soft target on the sampled states.
More concretely, we change $\pi_\mathcal{R}(s, \mathbf{y})$ into a policy $\pi_\mathcal{E}(s)$
which samples an action $a$ from $q(a\mid s)^{\frac{1}{T}}$, where $T$ is the temperature
that controls the sharpness of the distribution \cite{DBLP:journals/corr/HintonVD15}.
The algorithm is shown in Algorithm \ref{algo:distill_ref}.
Since such distillation generate training instances from
exploration, we name it as {\it distillation from exploration}.
Blocks connected by in solid blue lines in Figure \ref{fig:workflow} show
the workflow of our {\it distillation from exploration}.

On the sampled states, reference decision from $\pi_\mathcal{R}$ 
is usually non-trivial to achieve, which makes learning from NLL loss infeasible.
In Section \ref{sec:vani-exp}, we empirically show that fully distilling
from the soft target, i.e. setting $\alpha = 1$ in Equation \ref{eq:distill}, 
achieves comparable performance with that both
from distillation and NLL.

\subsection{Distillation from Both}\label{sec:distill_both}

Distillation from reference can encourage the model to predict
the action made by the reference policy and distillation from
exploration learns the model on arbitrary states.
They transfer the generalization ability of the ensemble from 
different aspects. Hopefully combining them
can further improve the performance.
In this paper, we combine distillation from reference and exploration
with the following manner:
we use $\pi_\mathcal{R}$ and $\pi_\mathcal{E}$ to generate a set of
training states.
Then, we learn $p(a \mid s)$ on the generated states.
If one state was generated by the reference policy,
we minimize the interpretation of distillation and NLL loss.
Otherwise, we minimize the distillation loss only.

\section{Experiments}\label{sec:vani-exp}
We perform experiments on two tasks: 
transition-based dependency parsing and neural machine translation.
Both these two tasks are converted to search-based structured prediction
as Section \ref{sec:sbsp}.

For the transition-based parsing, 
we use the stack-lstm parsing model 
proposed by \citet{dyer-EtAl:2015:ACL-IJCNLP}
to parameterize the classifier.\footnote{The code for parsing experiments is available at: \url{https://github.com/Oneplus/twpipe}.}
For the neural machine translation, 
we parameterize the classifier as an 
LSTM encoder-decoder model by following \citet{luong-pham-manning:2015:EMNLP}.\footnote{We based our NMT experiments on OpenNMT \citep{klein-EtAl:2017:ACL-2017-System-Demonstrations}. 
	The code for NMT experiments is available at: \url{https://github.com/Oneplus/OpenNMT-py}.}
We encourage the reader of this paper to refer corresponding papers for more
details.

\subsection{Settings}
\subsubsection{Transition-based Dependency Parsing}

We perform experiments on Penn Treebank (PTB) dataset
with standard data split (Section 2-21 for training, Section 22 for development,
and Section 23 for testing). Stanford dependencies are converted
from the original constituent trees using Stanford CoreNLP 3.3.0\footnote{\url{stanfordnlp.github.io/CoreNLP/history.html}} by
following \citet{dyer-EtAl:2015:ACL-IJCNLP}.
Automatic
part-of-speech tags are assigned by 10-way jackknifing whose accuracy
is 97.5\%.
Labeled attachment score (LAS) excluding punctuation
are used in evaluation.
For the other hyper-parameters, we use
the same settings as \citet{dyer-EtAl:2015:ACL-IJCNLP}.
The best iteration and $\alpha$ is determined on the development set.

\citet{reimers-gurevych:2017:EMNLP2017} and others have
pointed out that neural network training is 
nondeterministic and depends on the seed for the random
number generator. To control for this
effect, they suggest to report the average of $M$ differently-seeded runs.
In all our dependency parsing,
we set $n = 20$.

\subsubsection{Neural Machine Translation}
\begin{figure}[t]
	\includegraphics[width=\columnwidth]{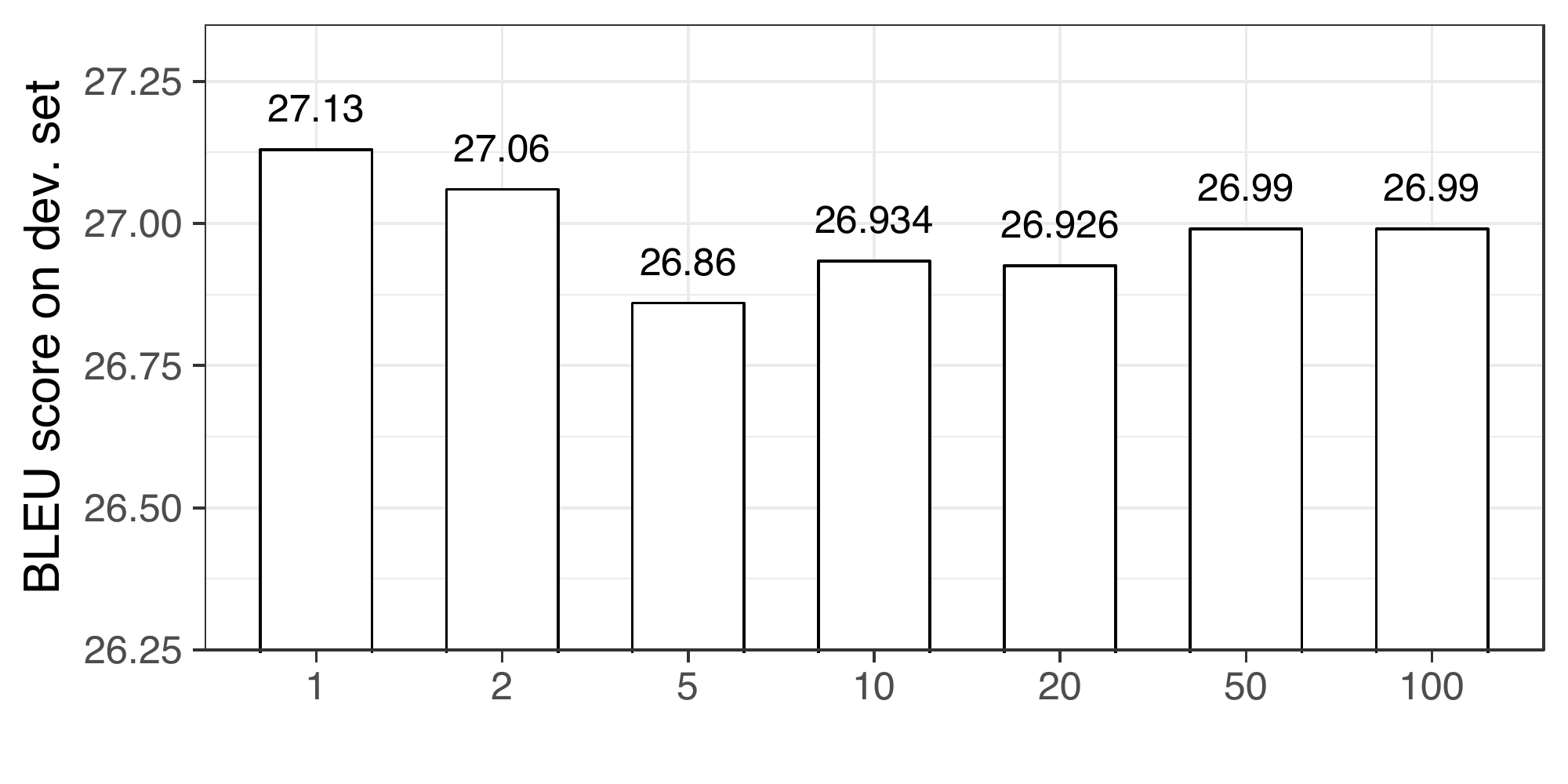}
	\caption{The effect of using different $K$s when approximating
	distillation loss with $K$-most probable actions in the machine translation experiments.}\label{fig:approx}
\end{figure}

We conduct our experiments on a small machine translation
dataset, which is the German-to-English portion of the 
IWSLT 2014 machine translation evaluation campaign.
The dataset contains around 153K training sentence pairs,
7K development sentence pairs, and 7K testing sentence pairs.
We use the same preprocessing as \citet{DBLP:journals/corr/RanzatoCAZ15}, 
which leads to a German vocabulary of about 30K entries and an
English vocabulary of 25K entries.
One-layer LSTM for both encoder and decoder with
256 hidden units are used by following \citet{wiseman-rush:2016:EMNLP2016}.
BLEU \citep{papineni-EtAl:2002:ACL} was used to evaluate the translator's
performance.\footnote{We use {\tt multi-bleu.perl} to evaluate our model's performance}
Like in the dependency parsing experiments, we run $M=10$
differently-seeded runs and report the averaged score.

Optimizing the distillation loss in Equation \ref{eq:distill} requires
enumerating over the action space. It is expensive for machine translation
since the size of the action space (vocabulary) is considerably large (25K in our experiments).
In this paper, we use the $K$-most probable actions (translations on target
side) on one state
to approximate the whole probability distribution of $q(a\mid s)$ as
$\sum_a q(a \mid s) \cdot \log p(a \mid s) \approx
\sum_k^K q(\hat{a}_k \mid s) \cdot \log p(\hat{a}_k \mid s)$, where
$\hat{a}_k$ is the $k$-th probable action.
We fix $\alpha$ to 1 and vary $K$ and evaluate the distillation model's
performance.
These results are shown in Figure \ref{fig:approx} where
there is no significant difference between different $K$s
and in speed consideration, we set $K$ to 1 in the following experiments.

\subsection{Results}\label{sec:exp-res}

\subsubsection{Transition-based Dependency Parsing}
\begin{table}[t]
	\centering
	\begin{tabular}{lc}
		\hline
		& LAS \\
		\hline
		Baseline & 90.83  \\
		Ensemble (20) & 92.73 \\
		Distill (reference, $\alpha$=1.0) & 91.99 \\
		Distill (exploration, $T$=1.0) & 92.00 \\
		Distill (both) & 92.14 \\
		\hdashline
		\citet{ballesteros-EtAl:2016:EMNLP2016}  (dyn. oracle) & 91.42 \\
		\citet{andor-EtAl:2016:P16-1} (local, B=1) & 91.02 \\
		\hdashline
		\citet{buckman-ballesteros-dyer:2016:EMNLP2016} (local, B=8) & 91.19 \\
		\citet{andor-EtAl:2016:P16-1} (local, B=32) & 91.70 \\
		\citet{andor-EtAl:2016:P16-1} (global, B=32) & 92.79 \\
		\citet{DBLP:journals/corr/DozatM16} & 94.08 \\
		\citet{kuncoro-16} & 92.06 \\
		\citet{kuncoro-17} & 94.60 \\
		\hline
	\end{tabular}
	\caption{The dependency parsing results. 
		Significance test \cite{NILSSON08.52} shows the improvement of our \textit{Distill (both)} over \textit{Baseline}
	is statistically significant with $p<0.01$.}\label{tbl:parse-res}
\end{table}

Table \ref{tbl:parse-res} shows our PTB experimental results.
From this result, we can see that the ensemble model outperforms
the baseline model by 1.90 in LAS. For our distillation from reference,
when setting $\alpha=1.0$, best performance on development set is achieved
and the test LAS is 91.99.

We tune the temperature $T$ during exploration and the results are shown in
Figure \ref{fig:temperature}. Sharpen the distribution during the sampling
process generally performs better on development set.
Our distillation from exploration model gets almost the same performance
as that from reference, but simply combing these two sets of data
outperform both models by achieving an LAS of 92.14.

We also compare our parser with the other parsers in Table \ref{tbl:parse-res}.
The second group shows the greedy transition-based parsers in previous literatures.
\citet{andor-EtAl:2016:P16-1} presented an alternative state
representation and explored both greedy and beam search decoding.
\cite{ballesteros-EtAl:2016:EMNLP2016} explores training the greedy parser with dynamic oracle.
Our distillation parser outperforms all these greedy counterparts.
The third group shows parsers trained on different techniques
including 
decoding with beam search \cite{buckman-ballesteros-dyer:2016:EMNLP2016,andor-EtAl:2016:P16-1},
training transition-based parser with beam search \citep{andor-EtAl:2016:P16-1},
graph-based parsing \citep{DBLP:journals/corr/DozatM16},
distilling a graph-based parser from the output of 20 parsers \cite{kuncoro-16},
and converting constituent parsing results to dependencies \cite{kuncoro-17}.
Our distillation parser still outperforms its transition-based counterparts but lags the others.
We attribute the gap between our parser with the other parsers to
the difference in parsing algorithms.

\subsubsection{Neural Machine Translation}

\begin{table}[t]
	\centering
	\begin{tabular}{lc}
		\hline
		& BLEU \\
		\hline
		Baseline & 22.79 \\
		Ensemble (10) & 26.26 \\
		Distill (reference, $\alpha$=0.8) & 24.76 \\
		Distill (exploration, $T$=0.1) & 24.64 \\
		Distill (both) & 25.44 \\
		\hdashline
		MIXER & 20.73 \\
		BSO (local, B=1) & 22.53 \\
		BSO (global, B=1) & 23.83 \\
		\hline
	\end{tabular}
	\caption{The machine translation results.
		MIXER denotes that of \citet{DBLP:journals/corr/RanzatoCAZ15},
		BSO  denotes that of \citet{wiseman-rush:2016:EMNLP2016}.
		Significance test \cite{koehn:2004:EMNLP} shows the improvement of our \textit{Distill (both)} over \textit{Baseline}
		is statistically significant with $p<0.01$.
	}\label{tbl:nmt-res}
\end{table}

\begin{figure}[t]
	\includegraphics[width=1\columnwidth]{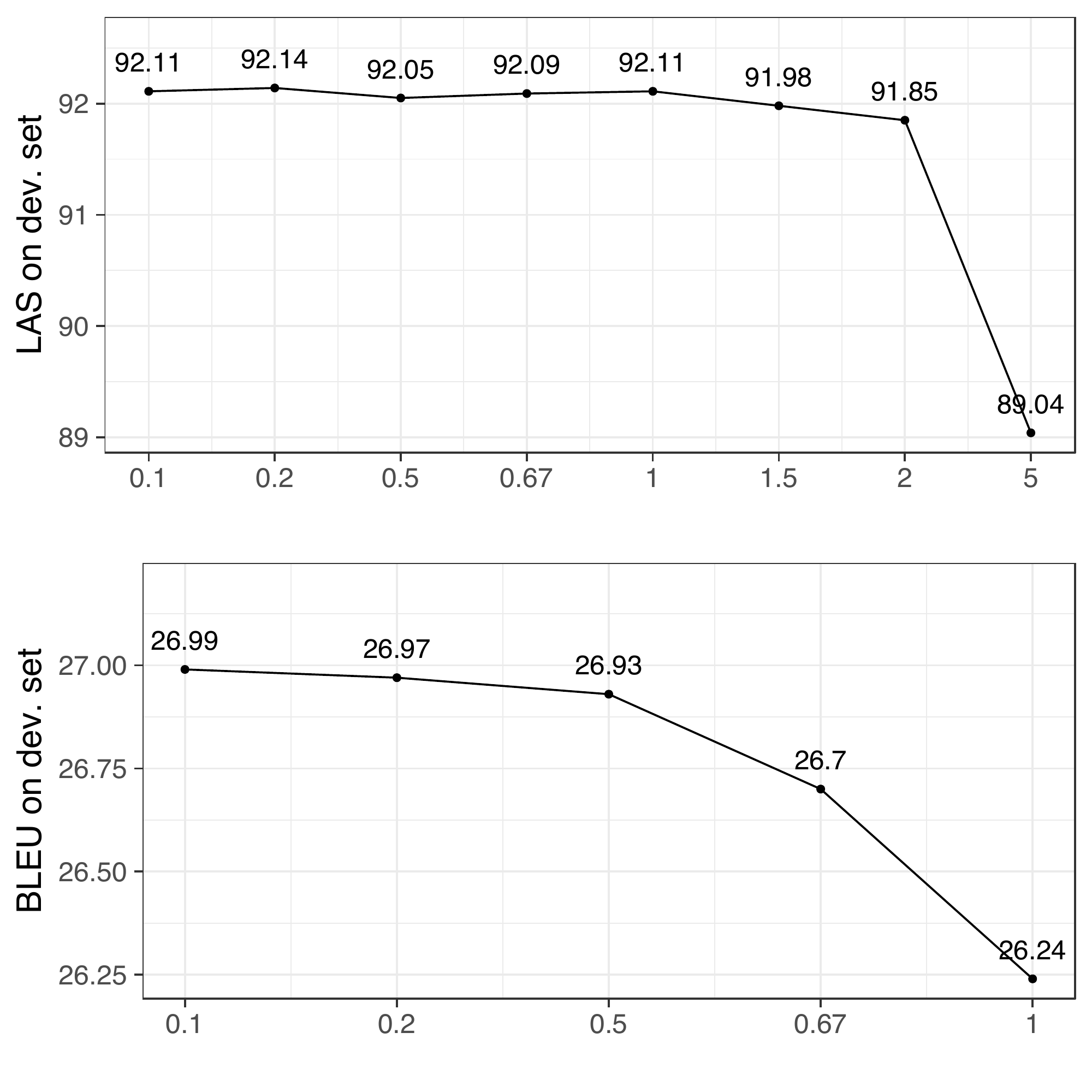}
	\caption{The effect of $T$ on PTB (above)
		 and IWSLT 2014 (below) development set.
	}\label{fig:temperature}
\end{figure}

Table \ref{tbl:nmt-res} shows the experimental results on IWSLT 2014 dataset.
Similar to the PTB parsing results, the ensemble 10 translators
outperforms the baseline translator by 3.47 in BLEU score.
Distilling from the ensemble by following the reference leads
to a single translator of 24.76 BLEU score.

Like in the parsing experiments, sharpen the distribution
when exploring the search space is more helpful to the model's
performance but the differences when $T\le0.2$ is not significant as shown in Figure \ref{fig:temperature}.
We set $T=0.1$ in our distillation from exploration experiments
since it achieves the best development score.
Table \ref{tbl:nmt-res} shows the exploration result of a BLEU score of 24.64 and
it slightly lags the best reference model.
Distilling from both the reference and exploration improves
the single model's performance by a large margin and achieves
a BLEU score of 25.44.

We also compare our model with other translation models including
the one trained with reinforcement learning \cite{DBLP:journals/corr/RanzatoCAZ15}
and that using beam search in training \cite{wiseman-rush:2016:EMNLP2016}.
Our distillation translator outperforms these models.

Both the parsing and machine translation experiments confirm that
it's feasible to distill a reasonable search-based structured prediction model
by just exploring the search space. Combining the reference and
exploration further improves the model's performance
and outperforms its greedy structured prediction counterparts.

%
%

\subsection{Analysis}\label{sec:analysis}

In Section \ref{sec:exp-res}, improvements from distilling the ensemble have been witnessed in
both the transition-based dependency parsing and neural machine translation experiments.
However, questions like ``Why the ensemble works better? 
Is it feasible to fully learn from the distillation loss without NLL?
Is learning from distillation loss stable?'' are yet to be answered.
In this section, we first study the ensemble's behavior on ``problematic'' states
to show its generalization ability. Then, we empirically study the feasibility
of fully learning from the distillation loss by studying the effect of $\alpha$ in the distillation
from reference setting. Finally, we show that learning
from distillation loss  is less sensitive to initialization and achieves a more stable model.

\subsubsection{Ensemble on ``Problematic'' States}\label{sec:ens-on-states}
\begin{table}
	\centering
	\begin{tabular}{l >{\centering\arraybackslash}p{5em} >{\centering\arraybackslash}p{5em}}
		\hline
		 & optimal-yet-ambiguous & non-optimal \\
		 \hline
		Baseline & 68.59 & 89.59\\
		Ensemble & 74.19 & 90.90 \\
		Distill (both) & 81.15 & 91.38 \\
		\hline
	\end{tabular}
\caption{The ranking performance of parsers' 
	output distributions evaluated in MAP on
	``problematic'' states.}\label{tbl:state-ana}
\end{table}
As mentioned in previous sections, ``problematic'' states
which is either ambiguous or non-optimal harm structured prediciton's 
performance.
Ensemble shows to improve the performance in Section \ref{sec:exp-res}, 
which indicates it does better on these states.
To empirically testify this, we use dependency parsing as a testbed
and study the ensemble's output distribution
using the dynamic oracle.

The dynamic oracle  \cite{goldberg-nivre:2012:PAPERS,TACL302} 
can be used to efficiently determine, given any state $s$, which
transition action leads to the best achievable parse from $s$; 
if some errors may have already made, what is the best the parser can do,
going forward?  This allows us to analyze the accuracy of each
parser's individual decisions, in the ``problematic'' states.
In this paper, we evaluate the output distributions of the
baseline and ensemble parser against
the {\it reference actions} suggested by the dynamic oracle.
Since dynamic oracle yields more than one reference actions
due to ambiguities and previous mistakes and the output distribution
can be treated as their scoring, we evaluate them
as a ranking problem. Intuitively, when multiple reference actions exist,
a good parser should push probability mass to these actions.
We draw problematic states by sampling from our baseline parser.
The comparison in Table \ref{tbl:state-ana} shows
that the ensemble model significantly outperforms the baseline on ambiguous
and non-optimal states.
This observation indicates the ensemble's output distribution is more
``informative'', thus generalizes well on problematic states and achieves
better performance.
We also observe that the distillation model perform better than both the baseline and ensemble.
We attribute this to the fact that the distillation model is learned from exploration.
\subsubsection{Effect of $\alpha$}
\begin{figure}[t]
	\centering
	\includegraphics[width=\columnwidth]{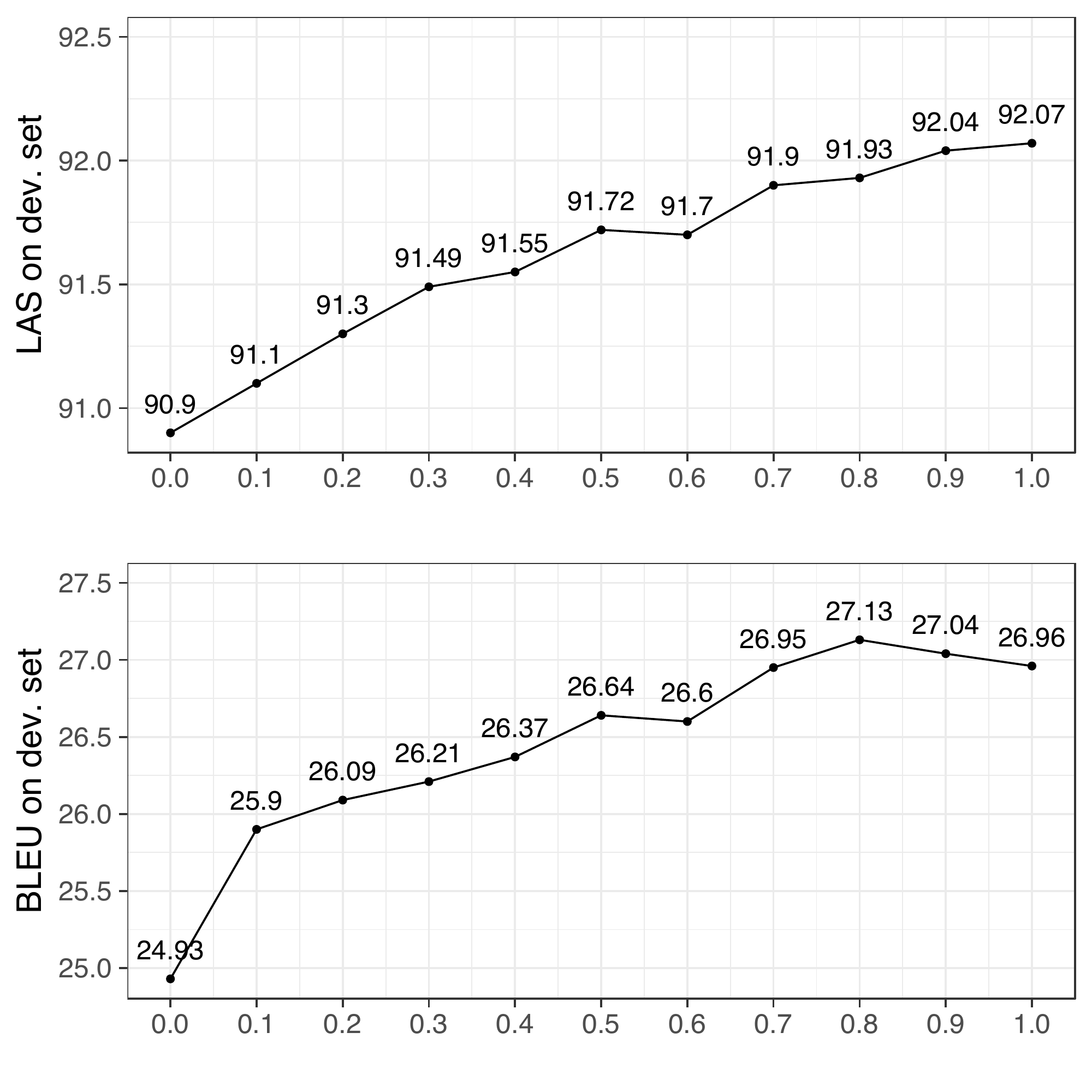}
	\caption{The effect of $\alpha$ on PTB (above)
		and IWSLT 2014 (below) development set.
	}\label{fig:alpha}
\end{figure}

Over our distillation from reference model, we study the effect of $\alpha$ in Equation \ref{eq:distill}.
We vary $\alpha$ from 0 to 1 by a step of 0.1 in both the transition-based dependency parsing
and neural machine translation experiments
and plot the model's performance on development sets in Figure \ref{fig:alpha}.
Similar trends are witnessed in both these two experiments that
model that's configured with larger $\alpha$ generally performs better than that with
smaller $\alpha$.
For the dependency parsing problem, the best development performance is achieved
when we set $\alpha=1$, and for the machine translation, the best $\alpha$ is 0.8.
There is only 0.2 point of difference between the best $\alpha$ model
and the one with $\alpha$ equals to 1.
Such observation indicates that
when distilling from the reference policy paying more attention
to the distillation loss rather than the NLL is more beneficial.
It also indicates that fully learning from 
the distillation loss outputted by the ensemble is reasonable because models configured with
$\alpha=1$ generally achieves good performance.

%
%
%
%

\subsubsection{Learning Stability}
\begin{figure}[t]
	\centering
	\begin{subfigure}[b]{0.49\columnwidth}
	\includegraphics[width=\columnwidth, trim={0, 1cm, 0, 1.5cm}, clip]{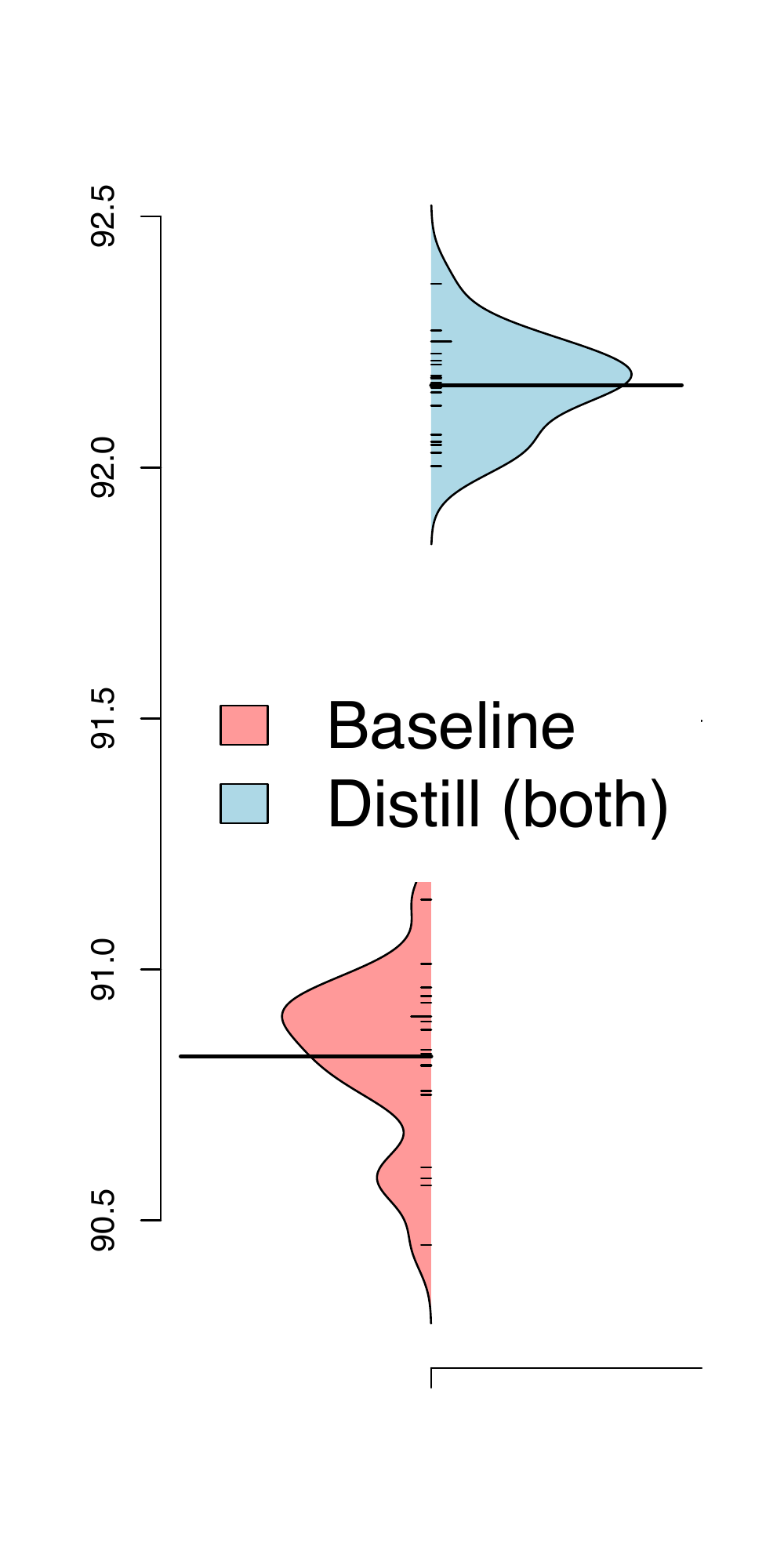}
	\end{subfigure}
	\begin{subfigure}[b]{0.49\columnwidth}
	\includegraphics[width=\columnwidth, trim={0, 1cm, 0, 1.5cm}, clip]{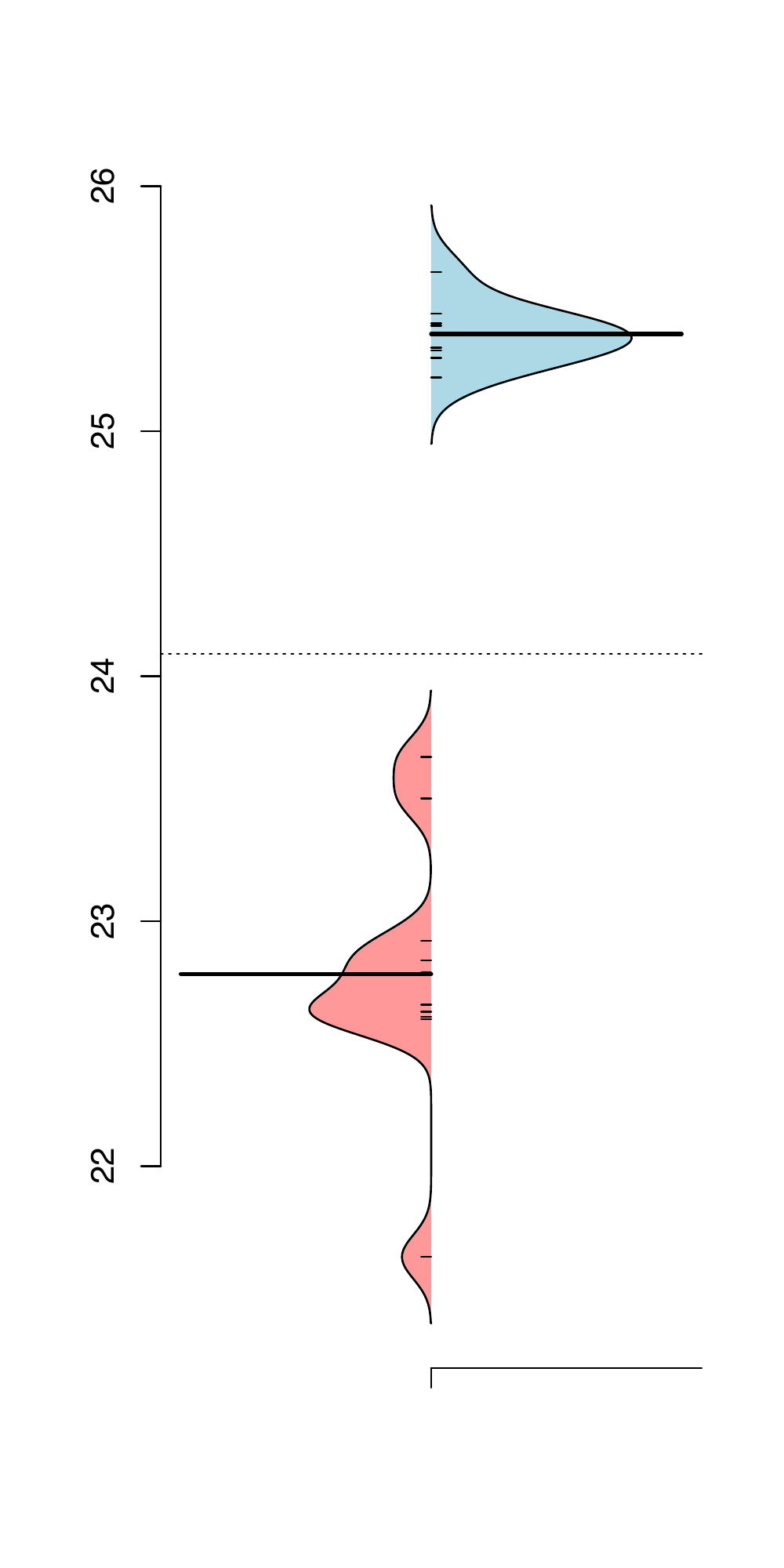}
	\end{subfigure}
	\caption{The distributions of scores for the baseline model and
		 our {\it distillation from both} on PTB test (left) and IWSLT 2014 test (right)
		 on differently-seeded runs.
	}\label{fig:stable}
\end{figure}

\begin{table}
	\centering
	\begin{tabular}{rcccc}
		\hline
		system & seeds & min & max & $\sigma$ \\
		\hline
		\multicolumn{2}{r}{\it PTB test} & & & \\
		Baseline & 20 & 90.45 & 91.14 & 0.17 \\
		Distill (both) & 20 & 92.00 & 92.37 & 0.09 \\
		\hline
		\hline
		\multicolumn{2}{r}{\it IWSLT 2014 test} & & & \\
		Baseline & 10 & 21.63 & 23.67 & 0.55 \\
		Distill (both) & 10 & 24.22 & 25.65 & 0.12 \\
		\hline
	\end{tabular}
\caption{The minimal, maximum, and standard derivation values
	on differently-seeded runs.}\label{tbl:stable}
\end{table}
Besides the improved performance,
knowledge distillation also leads to more stable learning.
The performance score distributions of differently-seed runs are depicted as
violin plot in Figure \ref{fig:stable}. Table \ref{tbl:stable}
also reveals the smaller standard derivations are achieved by
our distillation methods. As \citet{DBLP:journals/corr/KeskarMNST16} pointed out that
the generalization gap is not due to {\it overfit}, but due to
the network converge to {\it sharp minimizer} which generalizes worse,
we attribute the more stable training from our distillation model
as the distillation loss presents less {\it sharp minimizers}.

\section{Related Work}

Several works have been proposed to applying knowledge distillation
to NLP problems. \citet{kim-rush:2016:EMNLP2016} presented
a distillation model which focus on distilling the structured loss
from a large model into a small one which works on sequence-level.
In contrast to their work, we pay more attention to action-level distillation
and propose to do better action-level distillation by both from
reference and exploration.

\citet{DBLP:journals/corr/FreitagAS17} used an ensemble of 6-translators
to generate training reference.
Exploration was tried in their work with beam-search.
We differ their work by training the single model to match
the distribution of the ensemble.

Using ensemble in exploration was also studied in reinforcement learning
community \cite{NIPS2016_6501}.
In addition to distilling the ensemble on the labeled training data,
a line of semi-supervised learning works show that
it's effective to transfer knowledge of cumbersome model
into a simple one on the unlabeled data \cite{liang:icml08,li-zhang-chen:2014:P14-1}.
Their extensions to knowledge distillation call for further study.

\citet{kuncoro-16} proposed to compile the knowledge from
an ensemble of 20 transition-based parsers
into a voting and distill the knowledge
by introducing the voting results as a regularizer
in learning a graph-based parser.
Different from their work, we directly do the distillation on
the classifier of the transition-based parser.

Besides the attempts for directly using the  knowledge distillation technique,
\citet{stahlberg-byrne:2017:EMNLP2017} propose to first 
build the ensemble of several machine translators into one network by unfolding
and then use SVD to shrink its parameters, which can be treated as another kind of knowledge distillation.

\section{Conclusion}
In this paper, we study knowledge distillation for search-based structured prediction
and propose to distill an ensemble into a single model both from reference and exploration states.
Experiments on transition-based dependency parsing and machine translation
show that our distillation method significantly improves the single model's performance.
Comparison analysis gives empirically guarantee for our distillation method.

\section*{Acknowledgments}
We thank the anonymous reviewers for their helpful comments and suggestions.
This work was supported by the National Key Basic Research
Program of China via grant 2014CB340503 and the
National Natural Science Foundation of China (NSFC) via
grant 61632011 and 61772153.


\end{document}